\documentclass[10pt,twocolumn,letterpaper]{article}

\usepackage{cvpr}
\usepackage{times}
\usepackage{epsfig}
\usepackage{graphicx}
\usepackage{amsmath}
\usepackage{amssymb}

\usepackage{algorithm}
\usepackage{algpseudocode}% http://ctan.org/pkg/algorithmicx
\algrenewcommand\algorithmicindent{.9em}%

\usepackage{comment}
\usepackage[british,UKenglish,USenglish,english,american]{babel}
\usepackage{multirow}

\newcommand{\tabincell}[2]{\begin{tabular}{@{}#1@{}}#2\end{tabular}}

\usepackage[pagebackref=true,breaklinks=true,letterpaper=true,colorlinks,bookmarks=false]{hyperref}

\cvprfinalcopy % *** Uncomment this line for the final submission

\ifcvprfinal\fi
\begin{document}

%%%%%%%%% TITLE
\title{Deep Feature Flow for Video Recognition}

%\author{Xizhou Zhu\thanks{This work is done when Xizhou Zhu and Yuwen Xiong are interns at Microsoft Research Asia} \qquad\qquad Yuwen Xiong$^{*}$ \qquad\qquad Jifeng Dai \qquad\qquad Lu Yuan \qquad\qquad Yichen Wei \vspace{8pt}\\
%	Microsoft Research\\
%	{\tt\small \{v-xizzhu,v-yuxio,jifdai,luyuan,yichenw\}@microsoft.com}
%}
\author{Xizhou Zhu$^{1,2}$\thanks{This work is done when Xizhou Zhu and Yuwen Xiong are interns at Microsoft Research Asia} \qquad Yuwen Xiong$^{2}$$^{*}$ \qquad Jifeng Dai$^{2}$ \qquad Lu Yuan$^{2}$ \qquad Yichen Wei$^{2}$ \vspace{8pt}\\
$^{1}$University of Science and Technology of China \qquad $^{2}$Microsoft Research\qquad\qquad\\
\hspace{0.7in}{\tt\small ezra0408@mail.ustc.edu.cn} \qquad\qquad  {\tt\small \{v-yuxio,jifdai,luyuan,yichenw\}@microsoft.com} \qquad\\
}

\maketitle
\thispagestyle{empty}

%%%%%%%%% ABSTRACT
\begin{abstract}
Deep convolutional neutral networks have achieved great success on image recognition tasks. Yet, it is non-trivial to transfer the state-of-the-art image recognition networks to videos as per-frame evaluation is too slow and unaffordable. We present \emph{deep feature flow}, a fast and accurate framework for video recognition. It runs the expensive convolutional sub-network only on sparse key frames and propagates their deep feature maps to other frames via a flow field. It achieves significant speedup as flow computation is relatively fast. The end-to-end training of the whole architecture significantly boosts the recognition accuracy. Deep feature flow is flexible and general. It is validated on two video datasets on object detection and semantic segmentation. It significantly advances the practice of video recognition tasks. Code is released at \url{https://github.com/msracver/Deep-Feature-Flow}.
\end{abstract}

%%%%%%%%% BODY TEXT
\section{Introduction}
Recent years have witnessed significant success of deep convolutional neutral networks (CNNs) for image recognition, \eg, image classification~\cite{krizhevsky2012imagenet,simonyan2015very,szegedy2015going,he2016deep}, semantic segmentation~\cite{long2015fully,chen2015semantic,zheng2015conditional}, and object detection~\cite{girshick2014rich,he2014spatial,girshick2015fast,ren2015faster,dai2016rfcn,liu2016ssd}. With their success, the recognition tasks have been extended from image domain to video domain, such as semantic segmentation on Cityscapes dataset~\cite{cordts2016cityscapes}, and object detection on ImageNet VID dataset~\cite{olga2015imagenet}. Fast and accurate video recognition is crucial for high-value scenarios, \eg, autonomous driving and video surveillance. Nevertheless, applying existing image recognition networks on individual video frames introduces unaffordable computational cost for most applications.

It is widely recognized that image content varies slowly over video frames, especially the high level semantics~\cite{wiskott2002slow,zou2012deep,jayaraman2016slow}. This observation has been used as means of regularization in feature learning, considering videos as unsupervised data sources~\cite{wiskott2002slow,jayaraman2016slow}. Yet, such data redundancy and continuity can also be exploited to reduce the computation cost. This aspect, however, has received little attention for video recognition using CNNs in the literature.

Modern CNN architectures~\cite{simonyan2015very,szegedy2015going,he2016deep} share a common structure. Most layers are convolutional and account for the most computation. The intermediate convolutional feature maps have the same spatial extent of the input image (usually at a smaller resolution, \eg, $16\times$ smaller). They preserve the spatial correspondences between the low level image content and middle-to-high level semantic concepts~\cite{zeiler2014visualizing}. Such correspondence provides opportunities to cheaply propagate the features between nearby frames by spatial warping, similar to optical flow~\cite{horn1981determining}.

\begin{figure*}[t]
	\begin{center}
	\includegraphics[width=0.9\linewidth]{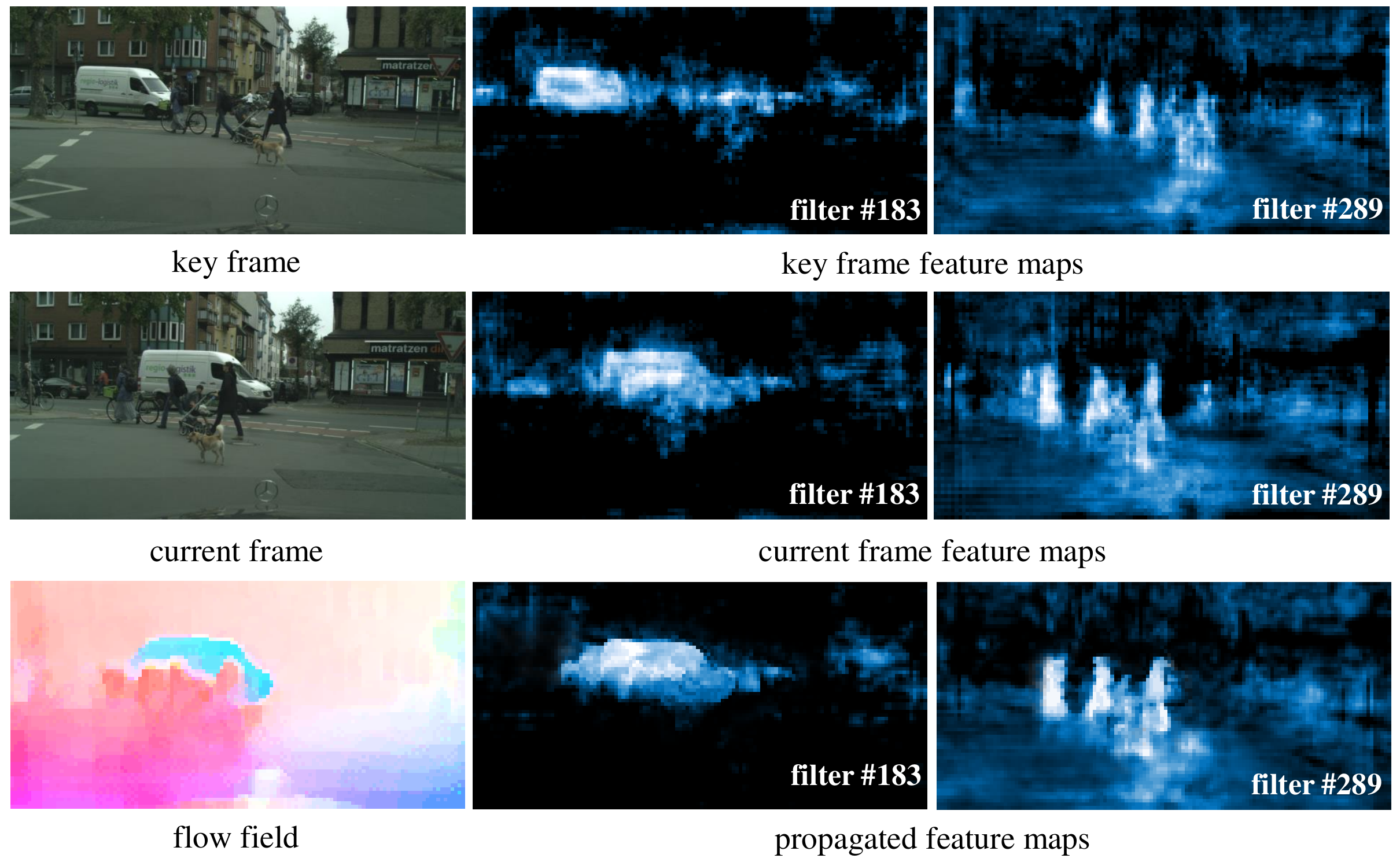}
	\end{center}\vspace{-1.0em}
	\caption{Motivation of proposed \emph{deep feature flow} approach. Here we visualize the two filters' feature maps on the last convolutional layer of our ResNet-101 model (see Sec.~\ref{sec.network_detail} for details). The convolutional feature maps are similar on two nearby frames. They can be cheaply propagated from the key frame to current frame via a flow field.}
	\label{fig.motivation}
\end{figure*}

In this work, we present \emph{deep feature flow}, a fast and accurate approach for video recognition. It applies an image recognition network on sparse key frames. It propagates the deep feature maps from key frames to other frames via a flow field. As exemplifed in Figure~\ref{fig.motivation}, two intermediate feature maps are responsive to ``car'' and ``person'' concepts. They are similar on two nearby frames. After propagation, the propagated features are similar to the original features.

Typically, the flow estimation and feature propagation are much faster than the computation of convolutional features. Thus, the computational bottleneck is avoided and significant speedup is achieved. When the flow field is also estimated by a network, the entire architecture is trained end-to-end, with both image recognition and flow networks optimized for the recognition task. The recognition accuracy is significantly boosted.

In sum, deep feature flow is a fast, accurate, general, and end-to-end framework for video recognition. It can adopt most state-of-the-art image recognition networks in the video domain. Up to our knowledge, it is the first work to jointly train flow and video recognition tasks in a deep learning framework. Extensive experiments verify its effectiveness on video object detection and semantic segmentation tasks, on recent large-scale video datasets. Compared to per-frame evaluation, our approach achieves unprecedented speed (up to $10\times$ faster, real time frame rate) with moderate accuracy loss (a few percent). The high performance facilitates video recognition tasks in practice. Code is released at \url{https://github.com/msracver/Deep-Feature-Flow}.

\section{Related Work}
To our best knowledge, our work is unique and there is no previous similar work to directly compare with. Nevertheless, it is related to previous works in several aspects, as discussed below.

\textbf{Image Recognition} Deep learning has been successful on image recognition tasks. The network architectures have evolved and become powerful on image classification~\cite{krizhevsky2012imagenet,simonyan2015very,szegedy2015going, he2015delving,ioffe2015batch,he2016deep}. For object detection, the region-based methods~\cite{girshick2014rich,he2014spatial,girshick2015fast,ren2015faster,dai2016rfcn} have become the dominant paradigm. For semantic segmentation, fully convolutional networks (FCNs)~\cite{long2015fully,chen2015semantic,zheng2015conditional} have dominated the field. However, it is computationally unaffordable to directly apply such image recognition networks on all the frames for video recognition. Our work provides an effective and efficient solution.

\textbf{Network Acceleration} Various approaches have been proposed to reduce the computation of networks. To name a few, in~\cite{zhang2015accelerating,girshick2015fast} matrix factorization is applied to decompose large network layers into multiple small layers. In~\cite{courbariaux2015binaryconnect,rastegari2016xnor,hubara2016quantized}, network weights are quantized. These techniques work on single images. They are generic and complementary to our approach.

\textbf{Optical Flow} It is a fundamental task in video analysis. The topic has been studied for decades and dominated by variational approaches~\cite{horn1981determining,brox2004high}, which mainly address small displacements~\cite{weickert2006survey}. The recent focus is on large displacements~\cite{brox2011large}, and combinatorial matching (e.g., DeepFlow~\cite{weinzaepfel2013deepflow}, EpicFlow~\cite{revaud2015epicflow}) has been integrated into the variational approach. These approaches are all hand-crafted.

Deep learning and semantic information have been exploited for optical flow recently. FlowNet~\cite{dosovitskiy2015flownet} firstly applies deep CNNs to directly estimate the motion and achieves good result. The network architecture is simplified in the recent Pyramid Network~\cite{ranjan2016optical}. Other works attempt to exploit semantic segmentation information to help optical flow estimation~\cite{laura2016optical,bai2016exploiting,hur2016joint}, \eg, providing specific constraints on the flow according to the category of the regions.

Optical flow information has been exploited to help vision tasks, such as pose estimation~\cite{pfister2015flowing}, frame prediction~\cite{patraucean2015spatio}, and attribute transfer~\cite{zhang2011discriminative}. This work exploits optical flow to speed up general video recognition tasks.

\textbf{Exploiting Temporal Information in Video Recognition} T-CNN~\cite{kang2016tcnn} incorporates temporal and contextual information from tubelets in videos. The dense 3D CRF~\cite{abhijit2016densecrf} proposes long-range spatial-temporal regularization in semantic video segmentation. STFCN~\cite{mohsen2016stfcn} considers a spatial-temporal FCN for semantic video segmentation. These works operate on volume data, show improved recognition accuracy but greatly increase the computational cost. By contrast, our approach seeks to reduce the computation by exploiting temporal coherence in the videos. The network still runs on single frames and is fast.

\textbf{Slow Feature Analysis} High level semantic concepts usually evolve slower than the low level image appearance in videos. The deep features are thus expected to vary smoothly on consecutive video frames. This observation has been used to regularize the feature learning in videos~\cite{wiskott2002slow,jayaraman2016slow,zou2012deep,zhang2012slow,sun2014dl}. We conjecture that our approach may also benefit from this fact.

\textbf{Clockwork Convnets~\cite{shelhamer2016clockwork}} It is the most related work to ours as it also disables certain layers in the network on certain video frames and reuses the previous features. It is, however, much simpler and less effective than our approach.

About speed up, Clockwork only saves the computation of some layers (e.g., $1/3$ or $2/3$) in some frames (e.g., every other frame). As seen later, our method saves that on most layers (task network has only 1 layer) in most frames (e.g., 9 out of 10 frames). Thus, our speedup ratio is much higher.

About accuracy, Clockwork does not exploit the correspondence between frames and simply copies features. It only reschedules the computation of inference in an off-the-shelf network and does not perform fine-tuning or re-training. Its accuracy drop is pretty noticeable at even small speed up. In Table 4 and 6 of their arxiv paper, at 77\% full runtime (thus 1.3 times faster), Mean IU drops from 31.1 to 26.4 on NYUD, from 70.0 to 64.0 on Youtube, from 65.9 to 63.3 on Pascal, and from 65.9 to 64.4 on Cityscapes. By contrast, we re-train a two-frame network with motion considered end-to-end. The accuracy drop is small, \eg., from 71.1 to 70.0 on Cityscape while being 3 times faster (Figure~\ref{fig.accuracy_speed_tradeoff}, bottom).

About generality, Clockwork is only applicable for semantic segmentation with FCN. Our approach transfers general image recognition networks to the video domain.

\setlength{\tabcolsep}{8pt}
\renewcommand{\arraystretch}{1.2}
\begin{table}[t]
\begin{center}
\begin{tabular}{l | l}
\hline
$k$ & key frame index       \\
$i$ & current frame index     \\
$r$ & per-frame computation cost ratio, Eq.~(\ref{eq.per_frame_ratio}) \\
$l$ & key frame duration length   \\
$s$ & overall speedup ratio, Eq.~(\ref{eq.speedup_ratio})  \\
\hline
$\mathbf{I}_i, \mathbf{I}_k$  & video frames \\
$\mathbf{y}_i, \mathbf{y}_k$  & recognition results \\
$\mathbf{f}_k$                & convolutional feature maps on key frame \\
$\mathbf{f}_i$                & propagated feature maps on current frame \\
$\mathbf{M}_{i\rightarrow k}$ & 2D flow field  \\
$\mathbf{p}$, $\mathbf{q}$    & 2D location           \\
$\mathbf{S}_{i\rightarrow k}$ & scale field  \\
\hline
$\mathcal{N}$        & image recognition network \\
$\mathcal{N}_{feat}$  & sub-network for feature extraction \\
$\mathcal{N}_{task}$ & sub-network for recognition result \\
$\mathcal{F}$        & flow estimation function      \\
$\mathcal{W}$        & feature propagation function, Eq.~(\ref{eq.propagate_function})   \\
\hline
\end{tabular}
\end{center}
\caption{Notations.}
\label{table.notation}
\end{table}

\section{Deep Feature Flow}
\begin{figure}[t]
\begin{center}
\includegraphics[width=1.0\linewidth]{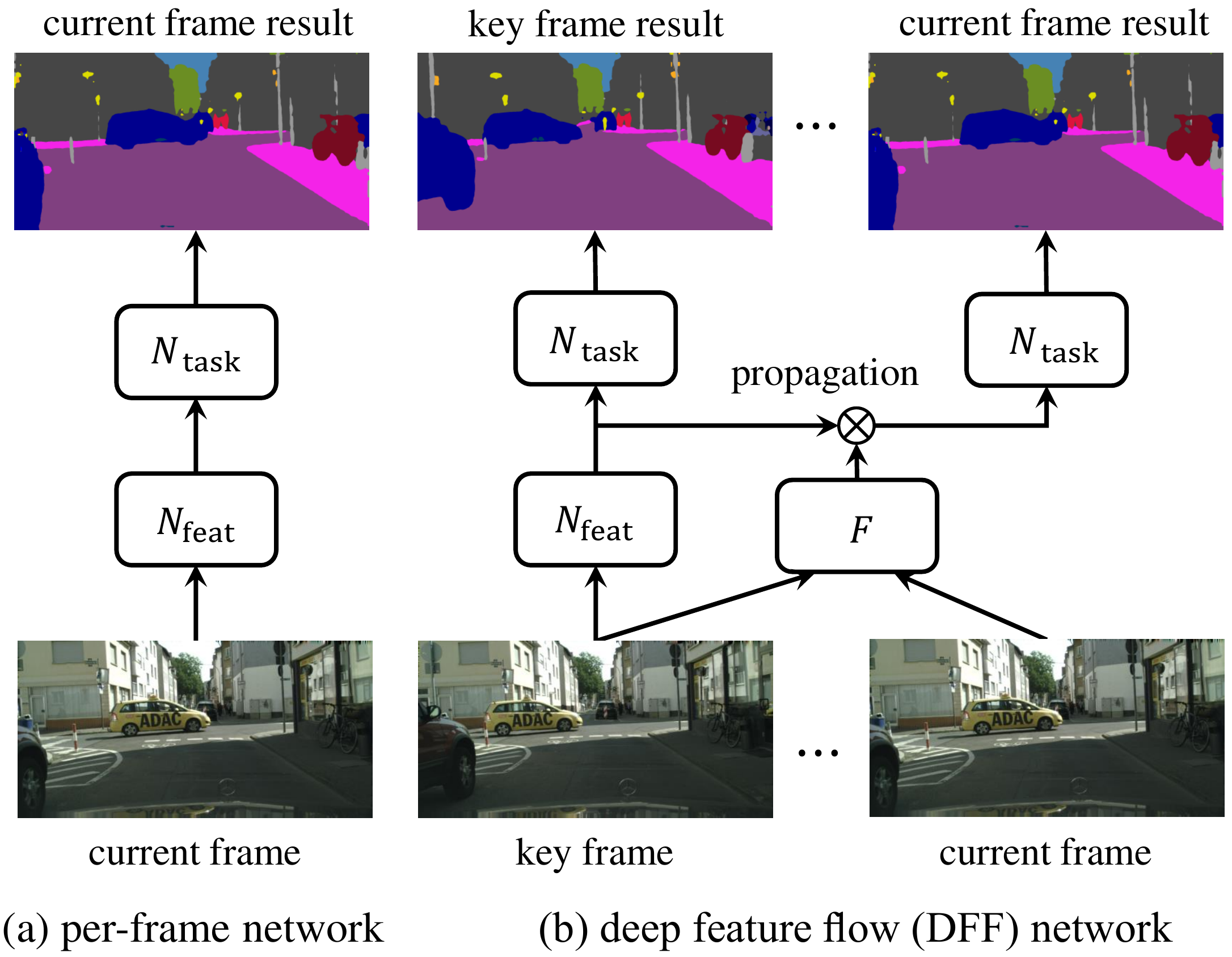}
\end{center}
   \caption{Illustration of video recognition using per-frame network evaluation (a) and the proposed deep feature flow (b).}
\label{fig.approach_overview}
\end{figure}

Table~\ref{table.notation} summarizes the notations used in this paper. Our approach is briefly illustrated in Figure~\ref{fig.approach_overview}.

\textbf{Deep Feature Flow Inference} Given an image recognition task and a feed-forward convolutional neutral network $\mathcal{N}$ that outputs result for input image $\mathbf{I}$ as $\mathbf{y}=\mathcal{N}(\mathbf{I})$. Our goal is to apply the network to all video frames $\mathbf{I}_i,i=0,...,\infty$, fast and accurately.

Following the modern CNN architectures~\cite{simonyan2015very,szegedy2015going,he2016deep} and applications~\cite{long2015fully,chen2015semantic,zheng2015conditional,girshick2014rich,he2014spatial,girshick2015fast,ren2015faster,dai2016rfcn}, without loss of generality, we
decompose $\mathcal{N}$ into two consecutive sub-networks. The first sub-network $\mathcal{N}_{feat}$, dubbed \emph{feature network}, is fully convolutional and outputs a number of intermediate feature maps, $\mathbf{f}=\mathcal{N}_{feat}(\mathbf{I})$. The second sub-network $\mathcal{N}_{task}$, dubbed \emph{task network}, has specific structures for the task and performs the recognition task over the feature maps, $\mathbf{y}=\mathcal{N}_{task}(\mathbf{f})$.

Consecutive video frames are highly similar. The similarity is even stronger in the deep feature maps, which encode high level semantic concepts~\cite{wiskott2002slow,jayaraman2016slow}. We exploit the similarity to reduce computational cost. Specifically, the feature network $\mathcal{N}_{feat}$ only runs on sparse key frames. The feature maps of a non-key frame $\mathbf{I}_i$ are propagated from its preceding key frame $\mathbf{I}_{k}$.

The features in the deep convolutional layers encode the semantic concepts and correspond to spatial locations in the image~\cite{zeiler2014visualizing}. Examples are illustrated in Figure~\ref{fig.motivation}. Such spatial correspondence allows us to cheaply propagate the feature maps by the manner of spatial warping.

Let $\mathbf{M}_{i\rightarrow k}$ be a two dimensional flow field. It is obtained by a flow estimation algorithm $\mathcal{F}$ such as~\cite{liu2008siftflow,dosovitskiy2015flownet}, $\mathbf{M}_{i\rightarrow k}=\mathcal{F}(\mathbf{I}_k,\mathbf{I}_i)$. It is bi-linearly resized to the same spatial resolution of the feature maps for propagation. It projects back a location $\mathbf{p}$ in current frame $i$ to the location $\mathbf{p}+\mathbf{\delta p}$ in key frame $k$, where $\mathbf{\delta p}=\mathbf{M}_{i\rightarrow k}(\mathbf{p})$.

As the values $\mathbf{\delta p}$ are in general fractional, the feature warping is implemented via bilinear interpolation
\begin{equation}
\mathbf{f}^{c}_{i}(\mathbf{p}) = \sum_{\mathbf{q}}G(\mathbf{q}, \mathbf{p}+\mathbf{\delta p})\mathbf{f}^{c}_{k}(\mathbf{q}),
\label{eq.feature_warp}
\end{equation}
where $c$ identifies a channel in the feature maps $\mathbf{f}$, $\mathbf{q}$ enumerates all spatial locations in the feature maps, and $G(\cdot, \cdot)$ denotes the bilinear interpolation kernel. Note that $G$ is two dimensional and is separated into two one dimensional kernels as

\begin{equation}
G(\mathbf{q}, \mathbf{p}+\mathbf{\delta p}) = g(q_x, p_x+\delta p_x)\cdot g(q_y, p_y+\delta p_y)
\label{eq.bilinear_interpolation_kernel},
\end{equation}
where $g(a, b) = max(0, 1-|a-b|)$.

We note that Eq.~\eqref{eq.feature_warp} is fast to compute as a few terms are non-zero.

The spatial warping may be inaccurate due to errors in flow estimation, object occlusion, etc. To better approximate the features, their amplitudes are modulated by a ``scale field" $\mathbf{S}_{i\rightarrow k}$, which is of the same spatial and channel dimensions as the feature maps. The scale field is obtained by applying a ``scale function" $\mathcal{S}$ on the two frames, $\mathbf{S}_{i\rightarrow k}=\mathcal{S}(\mathbf{I}_k,\mathbf{I}_i)$.

Finally, the feature propagation function is defined as
\begin{equation}
\mathbf{f}_i=\mathcal{W}(\mathbf{f}_k, \mathbf{M}_{i\rightarrow k}, \mathbf{S}_{i\rightarrow k}),
\label{eq.propagate_function}
\end{equation}
where $\mathcal{W}$ applies Eq.(\ref{eq.feature_warp}) for all locations and all channels in the feature maps, and multiples the features with scales $\mathbf{S}_{i\rightarrow k}$ in an element-wise way.

The proposed video recognition algorithm is called \emph{deep feature flow}. It is summarized in Algorithm~\ref{alg.DFF_inference}. Notice that any flow function $\mathcal{F}$, such as the hand-crafted low-level flow (e.g., SIFT-Flow~\cite{liu2008siftflow}), is readily applicable. Training the flow function
is not obligate, and the scale function $\mathcal{S}$ is set to
ones everywhere.

\begin{algorithm}[t]
\caption{Deep feature flow inference algorithm for video recognition.}
\begin{algorithmic}[1] % [1] for line numbers
\State \textbf{input}: video frames $\{\mathbf{I}_i\}$
\State $k=0$;                             \Comment{initialize key frame}
\State $\mathbf{f}_0 = \mathcal{N}_{feat}(\mathbf{I}_0)$
\State $\mathbf{y}_0 = \mathcal{N}_{task}(\mathbf{f}_0)$
\For{$i=1$ \textbf{to} $\infty$}
\If{$is\_key\_frame(i)$}                 \Comment{key frame scheduler}
\State $k=i$                             \Comment{update the key frame}
\State $\mathbf{f}_k = \mathcal{N}_{feat}(\mathbf{I}_k)$
\State $\mathbf{y}_k = \mathcal{N}_{task}(\mathbf{f}_k)$
\Else                                    \Comment{use feature flow}
\State $\mathbf{f}_i=\mathcal{W}(\mathbf{f}_k,\mathcal{F}(\mathbf{I}_k,\mathbf{I}_i),\mathcal{S}(\mathbf{I}_k,\mathbf{I}_i))$ \Comment{propagation}
\State $\mathbf{y}_i=\mathcal{N}_{task}(\mathbf{f}_i)$
\EndIf
\EndFor
\State \textbf{output}: recognition results $\{\mathbf{y}_{i}\}$
\end{algorithmic}
\label{alg.DFF_inference}
\end{algorithm}

\textbf{Deep Feature Flow Training} A flow function is originally designed to obtain correspondence of low-level image pixels. It can be fast in inference, but may not be accurate enough for the recognition task, in which the high-level feature maps change differently, usually slower than pixels~\cite{jayaraman2016slow,shelhamer2016clockwork}. To model such variations, we propose to also use a CNN to estimate the flow field and the scale field such that all the components can be jointly trained end-to-end for the task.

The architecture is illustrated in Figure~\ref{fig.approach_overview}(b). Training is performed by stochastic gradient descent (SGD). In each mini-batch, a pair of nearby video frames, $\{\mathbf{I}_k, \mathbf{I}_i\}$\footnote{The same notations are used for consistency although there is no longer the concept of ``key frame'' during training.}, $0\leq i-k\leq 9$, are randomly sampled. In the forward pass, feature network $\mathcal{N}_{feat}$ is applied on $\mathbf{I}_k$ to obtain the feature maps $\mathbf{f}_k$. Next, a \emph{flow network} $\mathcal{F}$ runs on the frames $\mathbf{I}_i$, $\mathbf{I}_k$ to estimate the flow field and the scale field. When $i>k$, feature maps $\mathbf{f}_k$ are propagated to $\mathbf{f}_i$ as in Eq.~(\ref{eq.propagate_function}). Otherwise, the feature maps are identical and no propagation is done. Finally, task network $\mathcal{N}_{task}$ is applied on $\mathbf{f}_i$ to produce the result $\mathbf{y}_i$, which incurs a loss against the ground truth result. The loss error gradients are back-propagated throughout to update all the components. Note that our training accommodates the special case when $i=k$ and degenerates to the per-frame training as in Figure~\ref{fig.approach_overview}(a).

The flow network is much faster than the feature network, as will be elaborated later. It is pre-trained on the Flying Chairs dataset~\cite{dosovitskiy2015flownet}. We then add the scale function $\mathcal{S}$ as a sibling output at the end of the network, by increasing the number of channels in the last convolutional layer appropriately. The scale function is initialized to all ones (weights and biases in the output layer are initialized as $0$s and $1$s, respectively). The augmented flow network is then fine-tuned as in Figure~\ref{fig.approach_overview}(b).

The feature propagation function in Eq.\eqref{eq.propagate_function} is unconventional. It is parameter free and fully differentiable. In back-propagation, we compute the derivative of the features in $\mathbf{f}_i$ with respect to the features in $\mathbf{f}_k$, the scale field $\mathbf{S}_{i\rightarrow k}$, and the flow field $\mathbf{M}_{i\rightarrow k}$. The first two are easy to compute using the chain rule. For the last, from Eq.~\eqref{eq.feature_warp} and~\eqref{eq.propagate_function}, for each channel $c$ and location $\mathbf{p}$ in current frame, we have
\begin{equation}
\frac{\partial \mathbf{f}_i^c(\mathbf{p})}{\partial \mathbf{M}_{i\rightarrow k}(\mathbf{p})} = \mathbf{S}^{c}_{i\rightarrow k}(\mathbf{p}) \sum_\mathbf{q} \frac{\partial G(\mathbf{q}, \mathbf{p}+\mathbf{\delta p})}{\partial \mathbf{\delta p}} \mathbf{f}_k^c(\mathbf{q}).
\label{eq.bilinear_derivative}
\end{equation}
The term $\frac{\partial G(\mathbf{q}, \mathbf{p}+\mathbf{\delta p})}{\partial \mathbf{\delta p}}$ can be derived from Eq.~\eqref{eq.bilinear_interpolation_kernel}. Note that the flow field $\mathbf{M}(\cdot)$ is two-dimensional and we use $\partial \mathbf{\delta p}$ to denote $\partial \delta p_x$ and $\partial \delta p_y$ for simplicity.

The proposed method can easily be trained on datasets where only sparse frames are annotated, which is usually the case due to the high labeling costs in video recognition tasks~\cite{silberman2012indoor,galasso2013unified,cordts2016cityscapes}. In this case, the per-frame training (Figure~\ref{fig.approach_overview}(a)) can only use annotated frames, while DFF can easily use all frames as long as frame $\mathbf{I}_i$ is annotated. In other words, DFF can fully use the data even with sparse ground truth annotation. This is potentially beneficial for many video recognition tasks.

\textbf{Inference Complexity Analysis} For each non-key frame, the computational cost ratio of the proposed approach (line 11-12 in Algorithm~\ref{alg.DFF_inference}) and per-frame approach (line 8-9) is
\begin{equation}
r=\frac{O(\mathcal{F})+O(\mathcal{S})+O(\mathcal{W})+O(\mathcal{N}_{task})}{O(\mathcal{N}_{feat})+O(\mathcal{N}_{task})},
\label{eq.per_frame_ratio}
\end{equation}
where $O(\cdot)$ measures the function complexity.

To understand this ratio, we firstly note that the complexity of $\mathcal{N}_{task}$ is usually small. Although its split point in $\mathcal{N}$ is kind of arbitrary, as verified in experiment, it is sufficient to keep only one learnable weight layer in $\mathcal{N}_{task}$ in our implementation (see Sec.~\ref{sec.network_detail}). While both $\mathcal{N}_{feat}$ and $\mathcal{F}$ have considerable complexity (Section~\ref{sec.network_detail}), we have $O(\mathcal{N}_{task})\ll O(\mathcal{N}_{feat})$ and $O(\mathcal{N}_{task})\ll O(\mathcal{F})$.

We also have $O(\mathcal{W})\ll O(\mathcal{F})$ and $O(\mathcal{S})\ll O(\mathcal{F})$ because $\mathcal{W}$ and $\mathcal{S}$ are very simple. Thus, the ratio in Eq.~\eqref{eq.per_frame_ratio} is approximated as

\begin{equation}
r\approx \frac{O(\mathcal{F})}{O(\mathcal{N}_{feat})}.
\label{eq.per_frame_ratio_appro}
\end{equation}

It is mostly determined by the complexity ratio of flow network $\mathcal{F}$ and feature network $\mathcal{N}_{feat}$, which can be precisely measured, \eg, by their FLOPs. Table~\ref{table.feat_flow_network_ratio} shows its typical values in our implementation.

Compared to the per-frame approach, the overall speedup factor in Algorithm~\ref{alg.DFF_inference} also depends on the sparsity of key frames. Let there be one key frame in every $l$ consecutive frames, the speedup factor is

\begin{equation}
s=\frac{l}{1+(l-1)*r}.
\label{eq.speedup_ratio}
\end{equation}

\textbf{Key Frame Scheduling} As indicated in Algorithm~\ref{alg.DFF_inference} (line 6) and Eq.~\eqref{eq.speedup_ratio}, a crucial factor for inference speed is when to allocate a new key frame.
In this work, we use a simple fixed key frame scheduling, that is, the key frame duration length $l$ is a fixed constant. It is easy to implement and tune. However, varied changes in image content may require a varying $l$ to provide a smooth tradeoff between accuracy and speed. Ideally, a new key frame should be allocated when the image content changes drastically. 

How to design effective and adaptive key frame scheduling can further improve our work. Currently it is beyond the scope of this work. Different video tasks may present different behaviors and requirements. Learning an adaptive key frame scheduler from data seems an attractive choice. This is worth further exploration and left as future work.

\setlength{\tabcolsep}{8pt}
\renewcommand{\arraystretch}{1.2}
\begin{table}
\begin{center}
\begin{tabular}{l | c | c | c}
\hline
           & \footnotesize FlowNet & \footnotesize FlowNet Half & \footnotesize FlowNet Inception \\
\hline
\footnotesize ResNet-50  &  9.20       &    33.56 &      68.97          \\
\hline
\footnotesize ResNet-101 &  12.71      &    46.30 &     95.24        \\
\hline
\end{tabular}
\end{center}
\caption{The approximated complexity ratio in Eq.~\eqref{eq.per_frame_ratio_appro} for different feature network $\mathcal{N}_{feat}$ and flow network $\mathcal{F}$, measured by their FLOPs. See Section~\ref{sec.network_detail}. Note that $r\ll 1$ and we use $\frac{1}{r}$ here for clarify. A significant per-frame speedup factor is obtained.}
\label{table.feat_flow_network_ratio}
\end{table}

\section{Network Architectures}
\label{sec.network_detail}
The proposed approach is general for different networks and recognition tasks. Towards a solid evaluation, we adopt the state-of-the-art architectures and important vision tasks.

\textbf{Flow Network} We adopt the state-of-the-art CNN based \emph{FlowNet} architecture (the ``Simple'' version)~\cite{dosovitskiy2015flownet} as default. We also designed two variants of lower complexity. The first one, dubbed \emph{FlowNet Half}, reduces the number of convolutional kernels in each layer of FlowNet by half and the complexity to $\frac{1}{4}$. The second one, dubbed \emph{FlowNet Inception}, adopts the Inception structure~\cite{szegedy2016rethinking} and reduces the complexity to $\frac{1}{8}$ of that of FlowNet. The architecture details are reported in Appendix~\ref{sec.flownet_inception}.

The three flow networks are pre-trained on the synthetic Flying Chairs dataset in~\cite{dosovitskiy2015flownet}. The output stride is 4. The input image is half-sized. The resolution of flow field is therefore $\frac{1}{8}$ of the original resolution. As the feature stride of the feature network is 16 (as described below), the flow field and the scale field is further down-sized by half using bilinear interpolation to match the resolution of feature maps. This bilinear interpolation is realized as a parameter-free layer in the network and also differentiated during training.

\textbf{Feature Network} We use ResNet models~\cite{he2016deep}, specifically, the ResNet-50 and ResNet-101 models pre-trained for ImageNet classification as default. The last 1000-way classification layer is discarded. The feature stride is reduced from 32 to 16 to produce denser feature maps, following the practice of DeepLab~\cite{chen2015semantic,chen2016deeplab} for semantic segmentation, and R-FCN~\cite{dai2016rfcn} for object detection. The first block of the conv5 layers are modified to have a stride of 1 instead of 2. The holing algorithm~\cite{chen2015semantic} is applied on all the $3\times3$ convolutional kernels in conv5 to keep the field of view (dilation=2). A randomly initialized $3\times3$ convolution is appended to conv5 to reduce the feature channel dimension to 1024, where the holing algorithm is also applied (dilation=6). The resulting 1024-dimensional feature maps are the intermediate feature maps for the subsequent task.

Table~\ref{table.feat_flow_network_ratio} presents the complexity ratio Eq.~\eqref{eq.per_frame_ratio_appro} of feature networks and flow networks.

\textbf{Semantic Segmentation} A randomly initialized $1\times1$ convolutional layer is applied on the intermediate feature maps to produce $(C+1)$ score maps, where $C$ is the number of categories and 1 is for background category. A following softmax layer outputs the per-pixel probabilities. Thus, the task network only has one learnable weight layer. The overall network architecture is similar to DeepLab with large field-of-view in~\cite{chen2016deeplab}.

\textbf{Object Detection} We adopt the state-of-the-art R-FCN~\cite{dai2016rfcn}. On the intermediate feature maps, two branches of fully convolutional networks are applied on the first half and the second half 512-dimensional of the intermediate feature maps separately, for sub-tasks of region proposal and detection, respectively.

In the region proposal branch, the RPN network~\cite{ren2015faster} is applied. We use $n_{\rm a}=9$ anchors (3 scales and 3 aspect ratios). Two sibling $1\times1$ convolutional layers output the $2n_{\rm a}$-dimensional objectness scores and the $4n_{\rm a}$-dimensional bounding box (bbox) regression values, respectively. Non-maximum suppression (NMS) is applied to generate 300 region proposals for each image. Intersection-over-union (IoU) threshold 0.7 is used.

In the detection branch, two sibling $1\times1$ convolutional layers output the position-sensitive score maps and bbox regression maps, respectively. They are of dimensions $(C+1)k^2$ and $4k^2$, respectively, where $k$ banks of classifiers/regressors are employed to encode the relative position information. See~\cite{dai2016rfcn} for details. On the position-sensitive score/bbox regression maps, position-sensitive ROI pooling is used to obtain the per-region classification score and bbox regression result. No free parameters are involved in the per-region computation. Finally, NMS is applied on the scored and regressed region proposals to produce the detection result, with IoU threshold 0.3.

\setlength{\tabcolsep}{5pt}
\renewcommand{\arraystretch}{1.2}
\begin{table*}[t]
	\begin{center}
		\begin{tabular}{ l | l | l }
			\hline
			method & training of image recognition network $\mathcal{N}$ & training of flow network $\mathcal{F}$ \\
			\hline
			\hline
			\emph{Frame} (oracle baseline)  & \footnotesize trained on single frames as in Fig.~\ref{fig.approach_overview} (a)                    & \footnotesize no flow network used      \\
			\hline
			\emph{SFF-slow}                   & \footnotesize same as \emph{Frame}        & \footnotesize SIFT-Flow~\cite{liu2008siftflow} (w/ best parameters), no training                        \\
			\emph{SFF-fast}                   & \footnotesize same as \emph{Frame}        & \footnotesize SIFT-Flow~\cite{liu2008siftflow} (w/ default parameters), no training                        \\
			\hline
			\emph{DFF}                   & \footnotesize trained on frame pairs as in Fig.~\ref{fig.approach_overview} (b)                  & \footnotesize init. on Flying Chairs~\cite{dosovitskiy2015flownet}, fine-tuned in Fig.~\ref{fig.approach_overview} (b)  \\
			\emph{DFF fix} $\mathcal{N}$ & \footnotesize same as \emph{Frame}, then fixed in Fig.~\ref{fig.approach_overview} (b) & \footnotesize same as \emph{DFF}                                                          \\
			\emph{DFF fix} $\mathcal{F}$ & \footnotesize same as \emph{DFF}                                                       & \footnotesize init. on Flying Chairs~~\cite{dosovitskiy2015flownet}, then fixed in Fig.~\ref{fig.approach_overview} (b) \\
			\emph{DFF separate}          & \footnotesize same as Frame       & \footnotesize init. on Flying Chairs~\cite{dosovitskiy2015flownet}       \\
			\hline
			
		\end{tabular}
	\end{center}
	\caption{Description of variants of deep feature flow (\emph{DFF}), shallow feature flow (\emph{SFF}), and the per-frame approach (Frame).}
	\label{table.algorithm_taxonomy}
\end{table*}

\section{Experiments}
Unlike image datasets, large scale video dataset is much harder to collect and annotate. Our approach is evaluated on the two recent datasets: Cityscapes~\cite{cordts2016cityscapes} for semantic segmentation, and ImageNet VID~\cite{olga2015imagenet} for object detection.

\subsection{Experiment Setup}

\textbf{Cityscapes} It is for urban scene understanding and autonomous driving. It contains snippets of street scenes collected from 50 different cities, at a frame rate of 17 fps. The train, validation, and test sets contain 2975, 500, and 1525 snippets, respectively. Each snippet has 30 frames, where the $20^{th}$ frame is annotated with pixel-level ground-truth labels for semantic segmentation. There are $30$ semantic categories. Following the protocol in \cite{chen2016deeplab}, training is performed on the train set and evaluation is performed on the validation set. The semantic segmentation accuracy is measured by the pixel-level mean intersection-over-union (mIoU) score.

In both training and inference, the images are resized to have shorter sides of 1024 and 512 pixels for the feature network and the flow network, respectively. In SGD training, 20K iterations are performed on 8 GPUs (each GPU holds one mini-batch, thus the effective batch size $\times8$), where the learning rates are $10^{-3}$ and $10^{-4}$ for the first 15K and the last 5K iterations, respectively.

\textbf{ImageNet VID} It is for object detection in videos. The training, validation, and test sets contain 3862, 555, and 937 fully-annotated video snippets, respectively. The frame rate is 25 or 30 fps for most snippets. There are 30 object categories, which are a subset of the categories in the ImageNet DET image dataset\footnote{http://www.image-net.org/challenges/LSVRC/}. Following the protocols in~\cite{kang2016tcnn,lee2016multi}, evaluation is performed on the validation set, using the standard mean average precision (mAP) metric.

In both training and inference, the images are resized to have shorter sides of 600 pixels and 300 pixels for the feature network and the flow network, respectively. In SGD training, 60K iterations are performed on 8 GPUs, where the learning rates are $10^{-3}$ and $10^{-4}$ for the first 40K and the last 20K iterations, respectively.

During training, besides the ImageNet VID train set, we also used the ImageNet DET train set (only the same 30 category labels are used), following the protocols in~\cite{kang2016tcnn,lee2016multi}. Each mini-batch samples images from either ImageNet VID or ImageNet DET datasets, at $2:1$ ratio.

\setlength{\tabcolsep}{3pt}
\renewcommand{\arraystretch}{1.2}
\begin{table}
	\begin{center}
		\begin{tabular}{ l | c | r | c | r}
			\hline
			\renewcommand{\arraystretch}{1.0} \multirow{2}{2cm}{Methods} & \multicolumn{2}{|c|}{\footnotesize{Cityscapes ($l$ = 5)}} & \multicolumn{2}{|c}{\footnotesize{ImageNet VID ($l$ = 10)}}\\
			\cline{2-5}
			& \footnotesize \tabincell{c}{mIoU(\%)} & \footnotesize \tabincell{c}{runtime (fps)} & \footnotesize \tabincell{c}{mAP(\%)} & \footnotesize \tabincell{c}{runtime (fps)} \\
			\hline
			\hline
			\emph{Frame}          &  \underline{71.1}  & 1.52\;\;\;\; &  \underline{73.9} & 4.05\;\;\;\; \\
			\hline
			\emph{SFF-slow}      &  67.8 & 0.08\;\;\;\;  &  70.7 & 0.26\;\;\;\; \\
			\emph{SFF-fast}      &  67.3 & 0.95\;\;\;\;  &  69.7 & 3.04\;\;\;\; \\
			\hline
			\emph{DFF}      &  \textbf{69.2} & 5.60\;\;\;\;  &  \textbf{73.1} & 20.25\;\;\;\; \\
			\emph{DFF fix $\mathcal{N}$}      &  68.8 & 5.60\;\;\;\;  &  72.3 & 20.25\;\;\;\; \\
			\emph{DFF fix $\mathcal{F}$}      &  67.0 & 5.60\;\;\;\;  &  68.8 & 20.25\;\;\;\; \\
			\emph{DFF separate}      &  66.9 & 5.60\;\;\;\;  &  67.4 & 20.25\;\;\;\; \\
			\hline
		\end{tabular}
	\end{center}
	\caption{Comparison of accuracy and runtime (mostly in GPU) of various approaches in Table~\ref{table.algorithm_taxonomy}. Note that, the runtime for \emph{SFF} consists of CPU runtime of SIFT-Flow and GPU runtime of \emph{Frame}, since SIFT-Flow only has CPU implementation.}
	\label{table.overall_accuracy_comparison}
\end{table}

\subsection{Evaluation Methodology and Results}
Deep feature flow is flexible and allows various design choices. We evaluate their effects comprehensively in the experiment. For clarify, we fix their default values throughout the experiments, unless specified otherwise. For feature network $\mathcal{N}_{feat}$, ResNet-101 model is default. For flow network $\mathcal{F}$, FlowNet (section~\ref{sec.network_detail}) is default. Key-frame duration length $l$ is 5 for Cityscapes~\cite{cordts2016cityscapes} segmentation and 10 for ImageNet VID~\cite{olga2015imagenet} detection by default, based on different frame rate of videos in the datasets..

For each snippet we evaluate $l$ image pairs, $(k, i)$, $k=i-l+1,...,i$, for each frame $i$ with ground truth annotation. Time evaluation is on a workstation with NVIDIA K40 GPU and Intel Core i7-4790 CPU.

\textbf{Validation of DFF Architecture} We compared DFF with several baselines and variants, as listed in Table~\ref{table.algorithm_taxonomy}.
\begin{itemize}
  \item \emph{Frame}: train $\mathcal{N}$ on single frames with ground truth.
  \item \emph{SFF}: use pre-computed large-displacement flow (e.g., SIFT-Flow~\cite{liu2008siftflow}). $\emph{SFF-fast}$ and $\emph{SFF-slow}$ adopt different parameters.
  \item \emph{DFF}: the proposed approach, $\mathcal{N}$ and $\mathcal{F}$ are trained end-to-end. Several variants include \emph{DFF fix} $\mathcal{N}$ (fix $\mathcal{N}$ in training), \emph{DFF fix} $\mathcal{F}$ (fix $\mathcal{F}$ in training), and \emph{DFF seperate} ($\mathcal{N}$ and $\mathcal{F}$ are separately trained).
\end{itemize}

Table~\ref{table.overall_accuracy_comparison} summarizes the accuracy and runtime of all approaches. We firstly note that the baseline \emph{Frame} is strong enough to serve as a reference for comparison. Our implementation resembles the state-of-the-art DeepLab~\cite{chen2016deeplab} for semantic segmentation and R-FCN~\cite{dai2016rfcn} for object detection. In DeepLab~\cite{chen2016deeplab}, an mIoU score of 69.2\% is reported with DeepLab large field-of-view model using ResNet-101 on Cityscapes validation dataset. Our \emph{Frame} baseline achieves slightly higher 71.1\%, based on the same ResNet model.

For object detection, \emph{Frame} baseline has mAP 73.9\% using R-FCN~\cite{dai2016rfcn} and ResNet-101. As a reference, a comparable mAP score of 73.8\% is reported in~\cite{kang2016tcnn}, by combining CRAFT~\cite{yang2016craft} and DeepID-net~\cite{ouyang2015deepid} object detectors trained on the ImageNet data, using both VGG-16~\cite{simonyan2015very} and GoogleNet-v2~\cite{ioffe2015batch} models, with various tricks (multi-scale training/testing, adding context information, model ensemble). We do not adopt above tricks as they complicate the comparison and obscure the conclusions.

\emph{SFF-fast} has a reasonable runtime but accuracy is significantly decreased. \emph{SFF-slow} uses the best parameters for flow estimation. It is much slower. Its accuracy is slightly improved but still poor. This indicates that an off-the-shelf flow may be insufficient.

The proposed \emph{DFF} approach has the best overall performance. Its accuracy is slightly lower than that of \emph{Frame} and it is 3.7 and 5.0 times faster for segmentation and detection, respectively. As expected, the three variants without using joint training have worse accuracy. Especially, the accuracy drop by fixing $\mathcal{F}$ is significant. This indicates a jointing end-to-end training (especially flow) is crucial.

We also tested another variant of \emph{DFF} with the scale function $\mathcal{S}$ removed (Algorithm~\ref{alg.DFF_inference}, Eq~\eqref{eq.propagate_function}, Eq.~\eqref{eq.bilinear_derivative}). The accuracy drops for both segmentation and detection (less than one percent). It shows that the scaled modulation of features is slightly helpful.

\textbf{Accuracy-Speedup Tradeoff} We investigate the tradeoff by varying the flow network $\mathcal{F}$, the feature network $\mathcal{N}_{feat}$, and key frame duration length $l$. Since Cityscapes and ImageNet VID datasets have different frame rates, we tested $l = 1, 2, ..., 10$ for segmentation and $l = 1, 2, ..., 20$ for detection.

\begin{figure}[t]
	\begin{center}
    \includegraphics[width=0.95\linewidth]{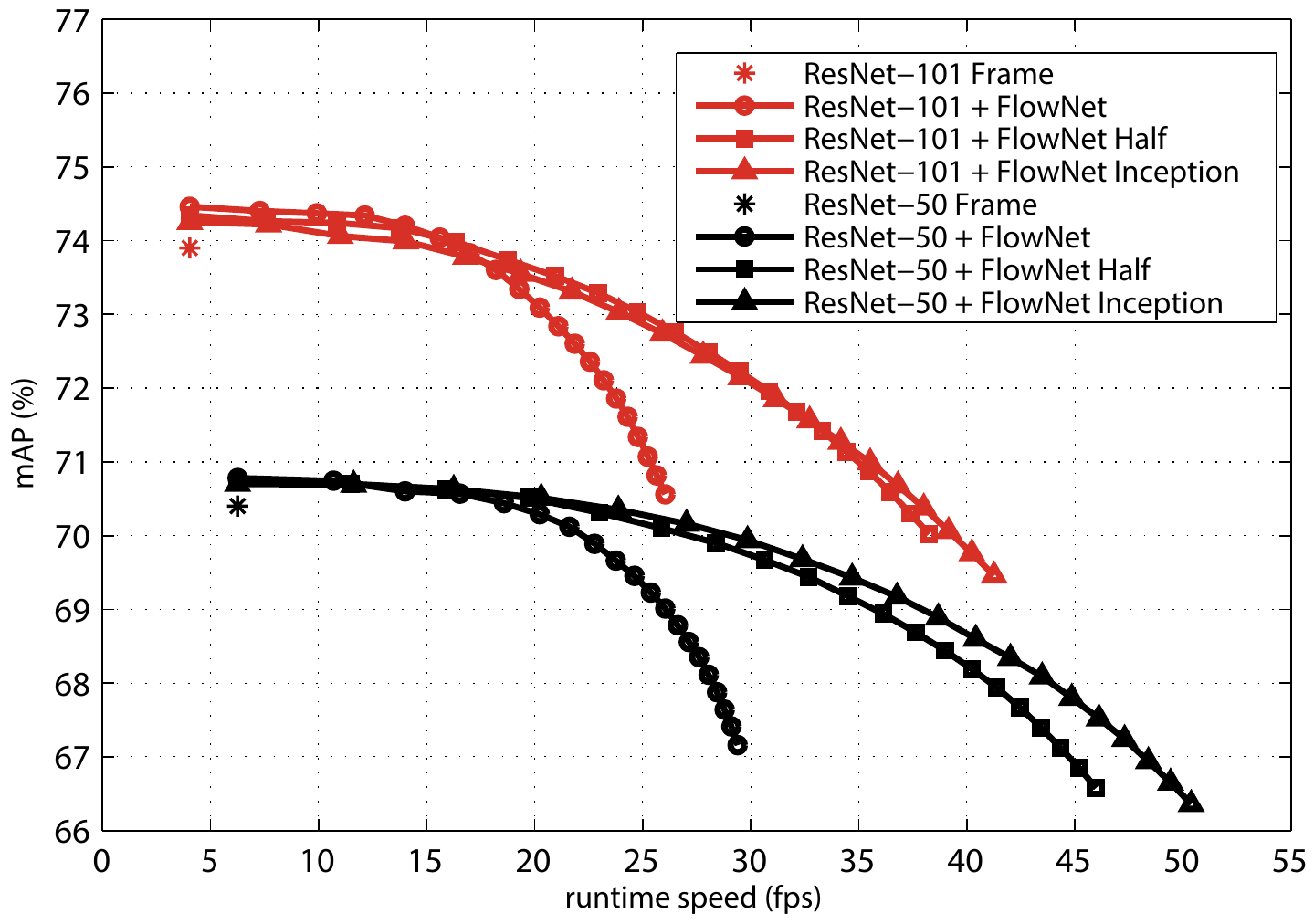}
    \includegraphics[width=0.95\linewidth]{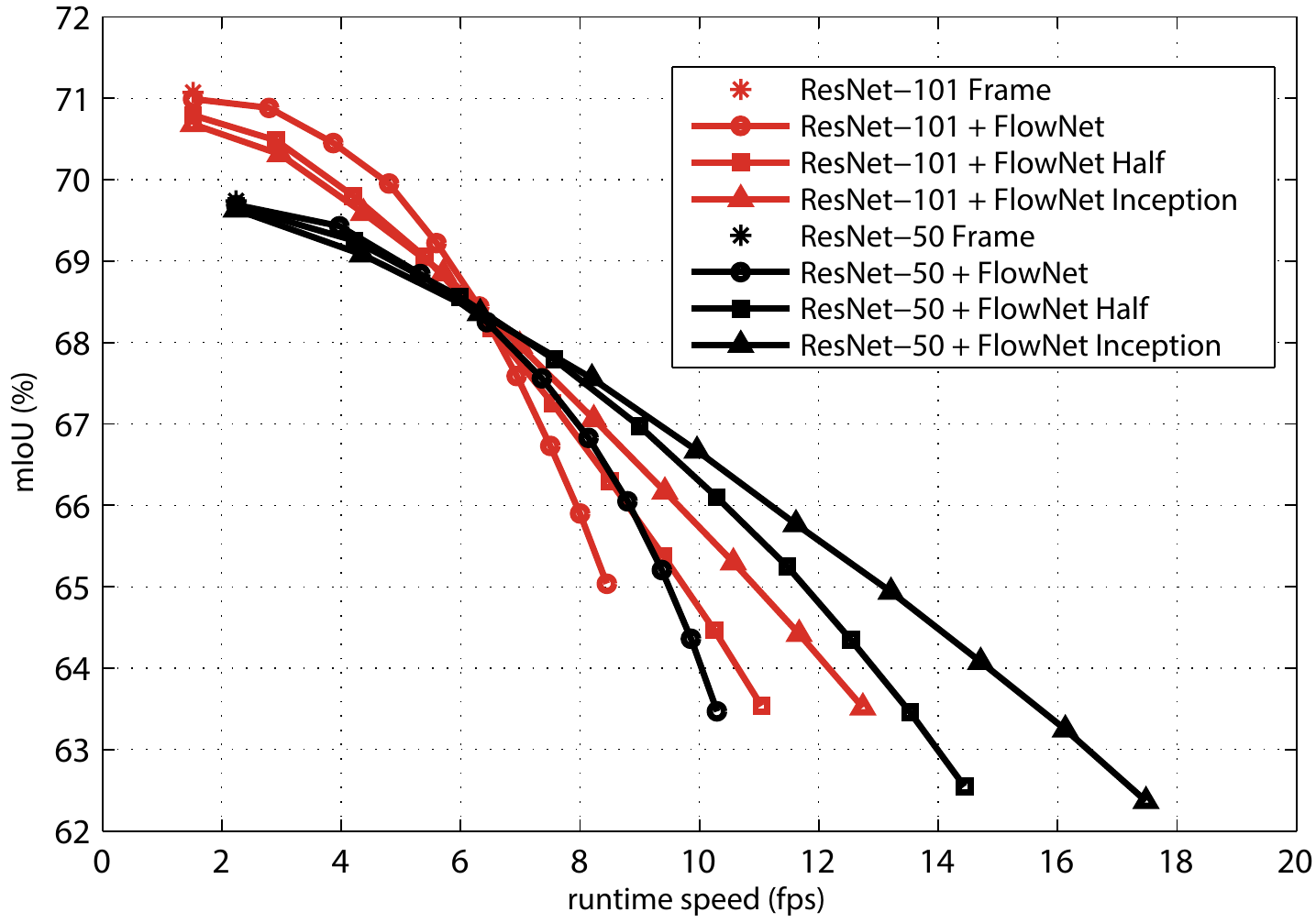}	
	\end{center}
	\caption{(better viewed in color) Illustration of accuracy-speed tradeoff under different implementation choices on ImageNet VID detection (\emph{top}) and on Cityscapes segmentation (\emph{bottom}).}
	\label{fig.accuracy_speed_tradeoff}
    \vspace{-1em}
\end{figure}

The results are summarized in Figure~\ref{fig.accuracy_speed_tradeoff}. Overall, DFF achieves significant speedup with decent accuracy drop. It smoothly trades in accuracy for speed and fits different application needs flexibly. For example, in detection, it improves 4.05 fps of \emph{ResNet-101 Frame} to 41.26 fps of \emph{ResNet-101 + FlowNet Inception}. The $10\times$ faster speed is at the cost of moderate accuracy drop from 73.9\% to 69.5\%. In segmentation, it improves 2.24 fps of \emph{ResNet-50 Frame} to 17.48 fps of \emph{ResNet-50 FlowNet Inception}, at the cost of accuracy drop from 69.7\% to 62.4\%.

What flow $\mathcal{F}$ should we use? From Figure~\ref{fig.accuracy_speed_tradeoff}, the smallest \emph{FlowNet Inception} is advantageous. It is faster than its two counterparts at the same accuracy level, most of the times.

What feature $\mathcal{N}_{feat}$ should we use? In high-accuracy zone, an accurate model \emph{ResNet-101} is clearly better than \emph{ResNet-50}. In high-speed zone, the conclusions are different on the two tasks. For detection, \emph{ResNet-101} is still advantageous. For segmentation, the performance curves intersect at around 6.35 fps point. For higher speed, \emph{ResNet-50} becomes better than \emph{ResNet-101}. The seemingly different conclusions can be partially attributed to the different video frame rates, the extents of dynamics on the two datasets. The Cityscapes dataset not only has a low frame rate 17 fps, but also more quick dynamics. It would be hard to utilize temporal redundancy for a long propagation. To achieve the same high speed, \emph{ResNet-101} needs a larger key frame length $l$ than \emph{ResNet-50}. This in turn significantly increases the difficulty of learning.

Above observations provide useful recommendations for practical applications. Yet, they are more heuristic than general, as they are observed only on the two tasks, on limited data. We plan to explore the design space more in the future.

\setlength{\tabcolsep}{3pt}
\renewcommand{\arraystretch}{1.2}
\begin{table}
	\begin{center}
		\small
		\begin{tabular}{l|c|c|c|c}
			\hline
			\footnotesize \renewcommand{\arraystretch}{1.0} \multirow{2}{2cm}{\footnotesize \tabincell{c}{\# layers in $\mathcal{N}_{task}$}} & \multicolumn{2}{|c|}{Cityscapes ($l$=5)} & \multicolumn{2}{|c}{ImageNet VID ($l$=10)}\\
			\cline{2-5}
			& \footnotesize mIoU(\%) & \footnotesize runtime (fps) & \footnotesize mAP(\%) & \footnotesize runtime (fps) \\
			\hline
			\hline
			\;\;\;\;\;21 & 69.1 & 2.87 & 73.2 & \;7.23 \\
			\;\;\;\;\;12 & 69.1 & 3.14 & 73.3 & \;8.04 \\
			\;\;\;\;\;5 & 69.2  & 3.89 & 73.2 & \;9.99 \\
			\;\;\;\;\;1 (default) & 69.2 & 5.60 & 73.1 & 20.25 \\
			\;\;\;\;\;0 & 69.5 & 5.61 & 72.7 & 20.40 \\
			\hline
		\end{tabular}
	\end{center}
	\caption{Results of using different split points for $\mathcal{N}_{task}$.}
	\label{table.split_point_exp}
\end{table}

\textbf{Split point of $\mathcal{N}_{task}$} Where should we split $\mathcal{N}_{task}$ in $\mathcal{N}$? Recall that the default $\mathcal{N}_{task}$ keeps one layer with learning weight (the $1\times 1$ conv over 1024-d feature maps, see Section~\ref{sec.network_detail}). Before this is the $3\times 3$ conv layer that reduces dimension to 1024. Before this is series of ``Bottleneck'' unit in ResNet~\cite{he2016deep}, each  consisting of 3 layers. We back move the split point to make different $\mathcal{N}_{task}$s with 5, 12, and 21 layers, respectively. The one with 5 layers adds the dimension reduction layer and one bottleneck unit (conv5c). The one with 12 layers adds two more units (conv5a and conv5b) at the beginning of conv5. The one with 21 layers adds three more units in conv4. We also move the only layer in default $\mathcal{N}_{task}$ into $\mathcal{N}_{feat}$, leaving $\mathcal{N}_{task}$ with 0 layer (with learnable weights). This is equivalent to directly propagate the parameter-free score maps, in both semantic segmentation and object detection.

Table~\ref{table.split_point_exp} summarizes the results. Overall, the accuracy variation is small enough to be neglected. The speed becomes lower when $\mathcal{N}_{task}$ has more layers. Using 0 layer is mostly equivalent to using 1 layer, in both accuracy and speed. We choose 1 layer as default as that leaves some tunable parameters after the feature propagation, which could be more general.

\begin{figure*}
\centering
\includegraphics[width=1.0\linewidth]{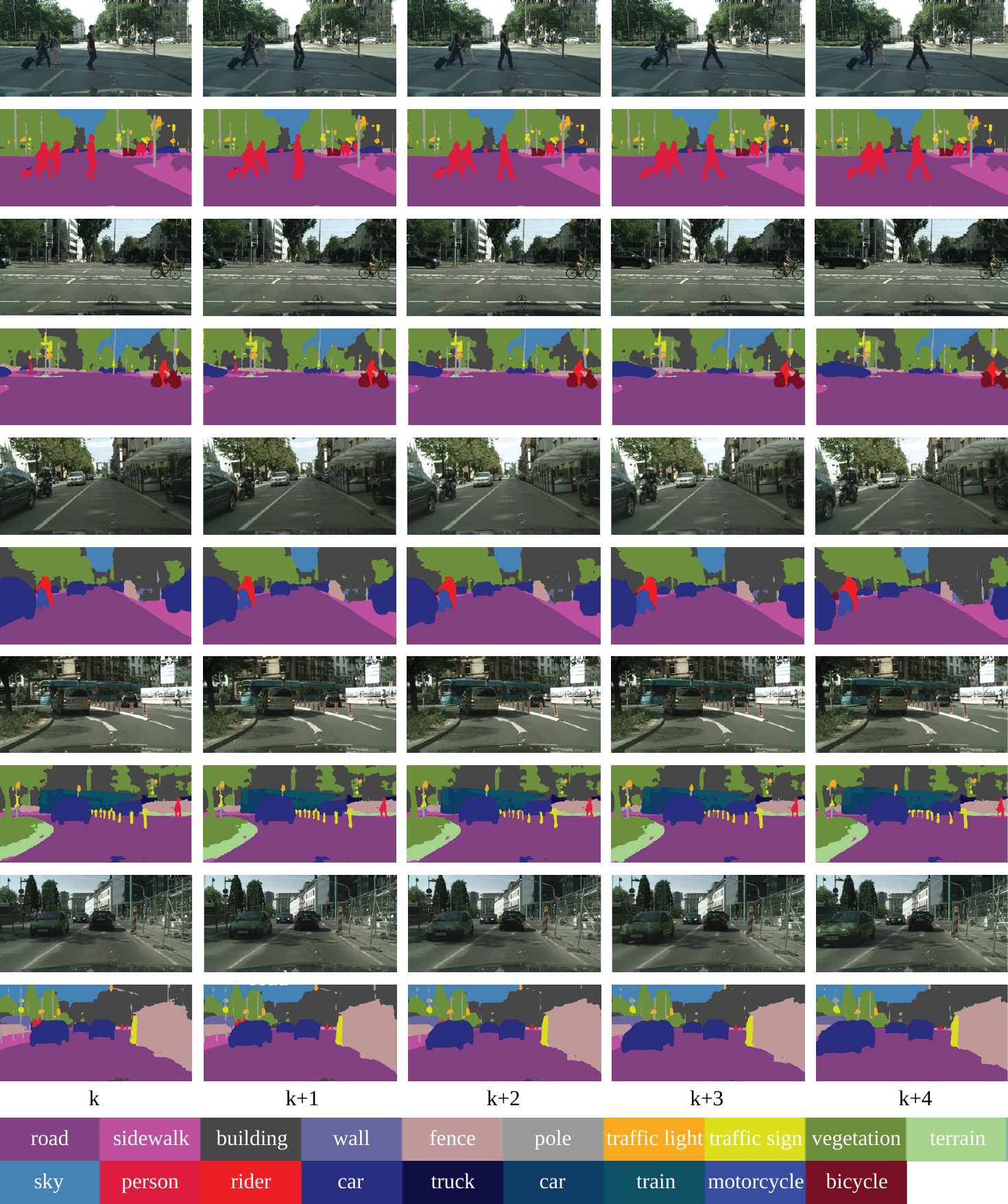}
\caption{Semantic segmentation results on Cityscapes validation dataset. The first column corresponds to the images and the results on the key frame (the $k^{th}$ frame). The following four columns correspond to the ${k+1}^{st}$, ${k+2}^{nd}$, ${k+3}^{rd}$ and ${k+4}^{th}$ frames, respectively.}
\label{fig.segmentation_result}
\end{figure*}

\begin{figure*}
\centering
\includegraphics[width=0.95\linewidth]{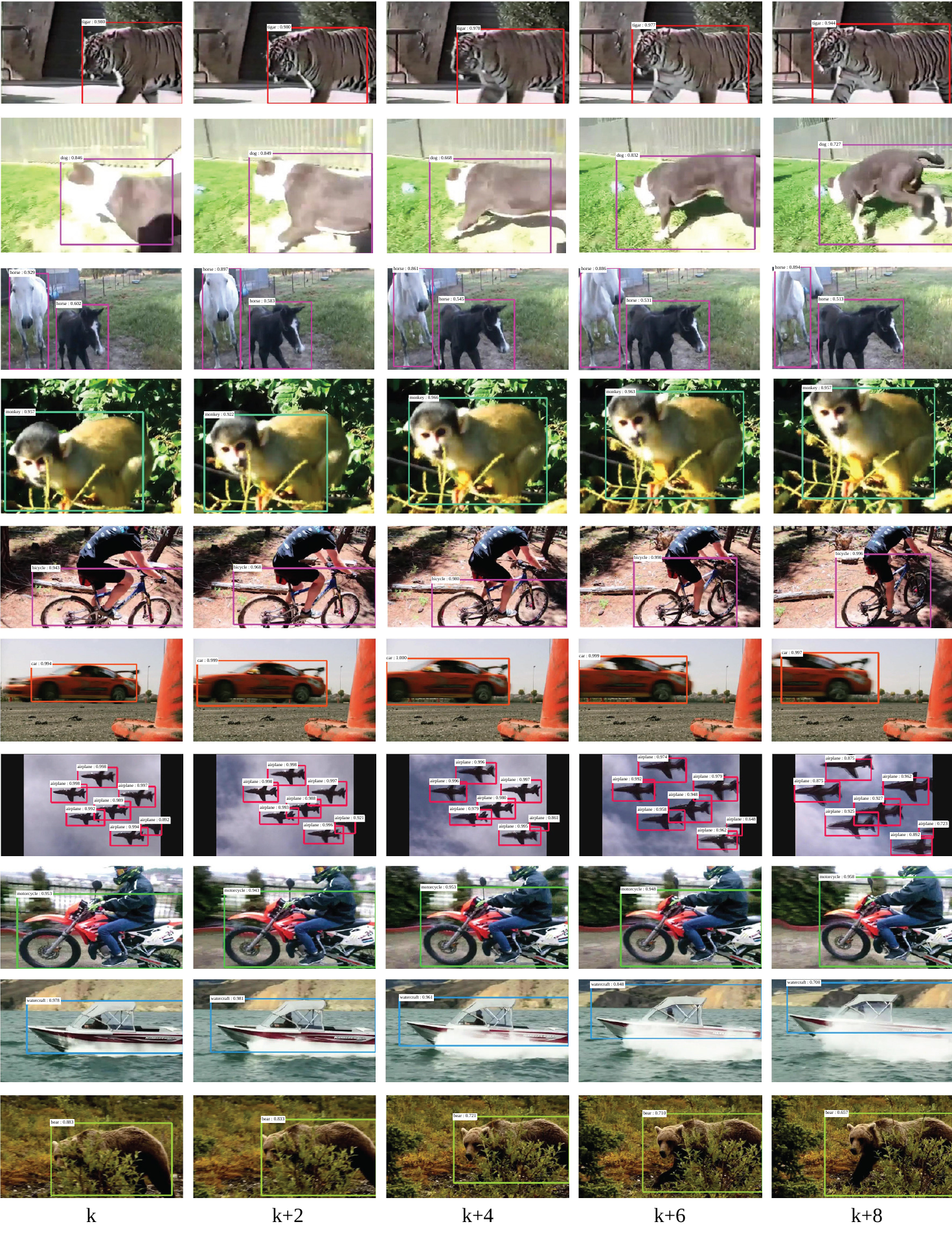}
\caption{Object detection results on ImageNet VID validation dataset. The first column corresponds to the images and the results on the key frame (the $k^{th}$ frame). The following four columns correspond to the ${k+2}^{nd}$, ${k+4}^{th}$, ${k+6}^{th}$ and ${k+8}^{th}$ frames, respectively.}
\label{fig.detection_result}
\end{figure*}

Example results of the proposed method are presented in Figure~\ref{fig.segmentation_result} and Figure~\ref{fig.detection_result} for video segmentation on CityScapes and video detection on ImageNet VID, respectively. More example results are available at \url{https://www.youtube.com/watch?v=J0rMHE6ehGw}.

\section{Future Work}

Several important aspects are left for further exploration. It would be interesting to exploit how the joint learning affects the flow quality. We are unable to evaluate as there lacks ground truth. Current optical flow works are also limited to either synthetic data~\cite{dosovitskiy2015flownet} or small real datasets, which is insufficient for deep learning.

Our method can further benefit from improvements in flow estimation and key frame scheduling. In this paper, we adopt FlowNet~\cite{dosovitskiy2015flownet} mainly because there are few choices. Designing faster and more accurate flow network will certainly receive more attention in the future. For key frame scheduling, a good scheduler may well significantly improve both speed and accuracy. And this problem is definitely worth further exploration.

We believe this work opens many new possibilities. We hope it will inspire more future work.

\appendix

\section{FlowNet Inception Architecture}
\label{sec.flownet_inception}

The architectures of FlowNet, FlowNet Half follow that of~\cite{dosovitskiy2015flownet} (the ``Simple" version), which are detailed in Table~\ref{table.flownet.original} and Table~\ref{table.flownet.half}, respectively. The architecture of  FlowNet Inception follows the design of the Inception structure~\cite{szegedy2016rethinking}, which is detailed in Table~\ref{table.flownet.inception}.

\setlength{\tabcolsep}{8pt}
\renewcommand{\arraystretch}{1.2}
\begin{table}[t]
	\begin{center}
		\begin{tabular}{|c|c|c|c|}
			\hline
			\multirow{2}[4]{*}{layer} & \multirow{2}[4]{*}{type} & \multirow{2}[4]{*}{stride} & \multirow{2}[4]{*}{\# output}\\
			&      &      &  \\
			\hline
			conv1 & 7x7 conv & 2    & 64 \\
			\hline
			conv2 & 5x5 conv & 2    & 128 \\
			\hline
			conv3 & 5x5 conv & 2    & 256 \\
			\hline
			conv3\_1 & 3x3 conv &      & 256 \\
			\hline
			conv4 & 3x3 conv & 2    & 512 \\
			\hline
			conv4\_1 & 3x3 conv &      & 512 \\
			\hline
			conv5 & 3x3 conv & 2    & 512 \\
			\hline
			conv5\_1 & 3x3 conv &      & 512 \\
			\hline
			conv6 & 3x3 conv & 2    & 1024 \\
			\hline
			conv6\_1 & 3x3 conv &      & 1024 \\
			\hline
		\end{tabular}	
	\end{center}
	\caption{The FlowNet network architecture.}
	\label{table.flownet.original}
\end{table}

\setlength{\tabcolsep}{8pt}
\renewcommand{\arraystretch}{1.2}
\begin{table}[t]
	\begin{center}
		\begin{tabular}{|c|c|c|c|}
			\hline
			\multirow{2}[4]{*}{layer} & \multirow{2}[4]{*}{type} & \multirow{2}[4]{*}{stride} & \multirow{2}[4]{*}{\# output}\\
			&      &      &  \\
			\hline
			conv1 & 7x7 conv & 2    & 32 \\
			\hline
			conv2 & 5x5 conv & 2    & 64 \\
			\hline
			conv3 & 5x5 conv & 2    & 128 \\
			\hline
			conv3\_1 & 3x3 conv &      & 128 \\
			\hline
			conv4 & 3x3 conv & 2    & 256 \\
			\hline
			conv4\_1 & 3x3 conv &      & 256 \\
			\hline
			conv5 & 3x3 conv & 2    & 256 \\
			\hline
			conv5\_1 & 3x3 conv &      & 256 \\
			\hline
			conv6 & 3x3 conv & 2    & 512 \\
			\hline
			conv6\_1 & 3x3 conv &      & 512 \\
			\hline
		\end{tabular}	
	\end{center}
	\caption{The FlowNet Half network architecture.}
	\label{table.flownet.half}
\end{table}

\setlength{\tabcolsep}{8pt}
\renewcommand{\arraystretch}{1.2}
\begin{table*}
	\begin{center}
		\begin{tabular}{|c|c|c|c|c|c|c|c|}
			\hline
			\multirow{2}[4]{*}{layer} & \multirow{2}[4]{*}{type} & \multirow{2}[4]{*}{stride} & \multirow{2}[4]{*}{\# output} & \multicolumn{4}{c|}{Inception/Reduction} \\
			\cline{5-8}     &      &      &      & \footnotesize \#1x1 & \footnotesize \#1x1-\#3x3 & \footnotesize \#1x1-\#3x3-\#3x3 & \footnotesize \#pool \\
			\hline
			conv1 & 7x7 conv & 2    & 32   &      &      &      &  \\
			\hline
			pool1 & 3x3 max pool & 2    & 32   &      &      &      &  \\
			\hline
			conv2 & Inception &      & 64   &      & 24-32 & 24-32-32 &  \\
			\hline
			conv3\_1 & 3x3 conv & 2    & 128  &      &      &      &  \\
			\hline
			conv3\_2 & Inception &      & 128  & 48   & 32-64 & 8-16-16 &  \\
			\hline
			conv3\_3 & Inception &      & 128  & 48   & 48-64 & 12-16-16 &  \\
			\hline
			conv4\_1 & Reduction & 2    & 256  & 32   & 112-128 & 28-32-32 & 64 \\
			\hline
			conv4\_2 & Inception &      & 256  & 96   & 112-128 & 28-32-32 &  \\
			\hline
			conv5\_1 & Reduction & 2    & 384  & 48   & 96-192 & 36-48-48 & 96 \\
			\hline
			conv5\_2 & Inception &      & 384  & 144  & 96-192 & 36-48-48 &  \\
			\hline
			conv6\_1 & Reduction & 2    & 512  & 64   & 192-256 & 48-64-64 & 128 \\
			\hline
			conv6\_2 & Inception &      & 512  & 192  & 192-256 & 48-64-64 &  \\
			\hline
		\end{tabular}	
	\end{center}
	\caption{The FlowNet Inception network architecture, following the design of the Inception structure~\cite{szegedy2016rethinking}. "Inception/Reduction" modules consist of four branches: 1x1 conv (\#1x1), 1x1 conv-3x3 conv (\#1x1-\#3x3), 1x1 conv-3x3 conv-3x3 conv (\#1x1-\#3x3-\#3x3), and 3x3 max pooling followed by 1x1 conv (\#pool, only for stride=2).}
	\label{table.flownet.inception}
\end{table*}

{\small
\bibliographystyle{ieee}
\bibliography{dff}

\begin{thebibliography}{10}\itemsep=-1pt

\bibitem{bai2016exploiting}
M.~Bai, W.~Luo, K.~Kundu, and R.~Urtasun.
\newblock Exploiting semantic information and deep matching for optical flow.
\newblock In {\em ECCV}, 2016.

\bibitem{brox2004high}
T.~Brox, A.~Bruhn, N.~Papenberg, and J.~Weickert.
\newblock High accuracy optical flow estimation based on a theory for warping.
\newblock In {\em ECCV}, 2004.

\bibitem{brox2011large}
T.~Brox and J.~Malik.
\newblock Large displacement optical flow: Descriptor matching in variational
  motion estimation.
\newblock {\em TPAMI}, 2011.

\bibitem{chen2015semantic}
L.-C. Chen, G.~Papandreou, I.~Kokkinos, K.~Murphy, and A.~L. Yuille.
\newblock Semantic image segmentation with deep convolutional nets and fully
  connected crfs.
\newblock In {\em ICLR}, 2015.

\bibitem{chen2016deeplab}
L.-C. Chen, G.~Papandreou, I.~Kokkinos, K.~Murphy, and A.~L. Yuille.
\newblock Deeplab: Semantic image segmentation with deep convolutional nets,
  atrous convolution, and fully connected crfs.
\newblock {\em arXiv preprint}, 2016.

\bibitem{cordts2016cityscapes}
M.~Cordts, M.~Omran, S.~Ramos, T.~Rehfeld, M.~Enzweiler, R.~Benenson,
  U.~Franke, S.~Roth, and B.~Schiele.
\newblock The cityscapes dataset for semantic urban scene understanding.
\newblock In {\em CVPR}, 2016.

\bibitem{courbariaux2015binaryconnect}
M.~Courbariaux, Y.~Bengio, and J.-P. David.
\newblock Binaryconnect: Training deep neural networks with binary weights
  during propagations.
\newblock In {\em NIPS}, 2015.

\bibitem{dai2016rfcn}
J.~Dai, Y.~Li, K.~He, and J.~Sun.
\newblock R-fcn: Object detection via region-based fully convolutional
  networks.
\newblock In {\em NIPS}, 2016.

\bibitem{dosovitskiy2015flownet}
A.~Dosovitskiy, P.~Fischer, E.~Ilg, P.~Hausser, C.~Hazirbas, and V.~Golkov.
\newblock Flownet: Learning optical flow with convolutional networks.
\newblock In {\em ICCV}, 2015.

\bibitem{mohsen2016stfcn}
M.~Fayyaz, M.~H. Saffar, M.~Sabokrou, M.~Fathy, and R.~Klette.
\newblock {STFCN:} spatio-temporal {FCN} for semantic video segmentation.
\newblock {\em arXiv preprint}, 2016.

\bibitem{galasso2013unified}
F.~Galasso, N.~Shankar~Nagaraja, T.~Jimenez~Cardenas, T.~Brox, and B.~Schiele.
\newblock A unified video segmentation benchmark: Annotation, metrics and
  analysis.
\newblock In {\em ICCV}, 2013.

\bibitem{girshick2015fast}
R.~Girshick.
\newblock {Fast R-CNN}.
\newblock In {\em ICCV}, 2015.

\bibitem{girshick2014rich}
R.~Girshick, J.~Donahue, T.~Darrell, and J.~Malik.
\newblock Rich feature hierarchies for accurate object detection and semantic
  segmentation.
\newblock In {\em CVPR}, 2014.

\bibitem{he2014spatial}
K.~He, X.~Zhang, S.~Ren, and J.~Sun.
\newblock Spatial pyramid pooling in deep convolutional networks for visual
  recognition.
\newblock In {\em ECCV}, 2014.

\bibitem{he2015delving}
K.~He, X.~Zhang, S.~Ren, and J.~Sun.
\newblock Delving deep into rectifiers: Surpassing human-level performance on
  imagenet classification.
\newblock In {\em ICCV}, 2015.

\bibitem{he2016deep}
K.~He, X.~Zhang, S.~Ren, and J.~Sun.
\newblock Deep residual learning for image recognition.
\newblock In {\em CVPR}, 2016.

\bibitem{horn1981determining}
B.~K. Horn and B.~G. Schunck.
\newblock Determining optical flow.
\newblock {\em Artificial intelligence}, 1981.

\bibitem{hubara2016quantized}
I.~{Hubara}, M.~{Courbariaux}, D.~{Soudry}, R.~{El-Yaniv}, and Y.~{Bengio}.
\newblock {Quantized Neural Networks: Training Neural Networks with Low
  Precision Weights and Activations}.
\newblock {\em arXiv preprint}, 2016.

\bibitem{hur2016joint}
J.~Hur and S.~Roth.
\newblock Joint optical flow and temporally consistent semantic segmentation.
\newblock In {\em ECCV CVRSUAD Workshop}, 2016.

\bibitem{ioffe2015batch}
S.~Ioffe and C.~Szegedy.
\newblock Batch normalization: Accelerating deep network training by reducing
  internal covariate shift.
\newblock In {\em ICML}, 2015.

\bibitem{jayaraman2016slow}
D.~Jayaraman and K.~Grauman.
\newblock Slow and steady feature analysis: higher order temporal coherence in
  video.
\newblock In {\em CVPR}, 2016.

\bibitem{kang2016tcnn}
K.~Kang, H.~Li, J.~Yan, X.~Zeng, B.~Yang, T.~Xiao, C.~Zhang, Z.~Wang, R.~Wang,
  and X.~Wang.
\newblock T-cnn: Tubelets with convolutional neural networks for object
  detection from videos.
\newblock In {\em CVPR}, 2016.

\bibitem{krizhevsky2012imagenet}
A.~Krizhevsky, I.~Sutskever, and G.~E. Hinton.
\newblock Imagenet classification with deep convolutional neural networks.
\newblock In {\em NIPS}, 2012.

\bibitem{abhijit2016densecrf}
A.~Kundu, V.~Vineet, and V.~Koltun.
\newblock Feature space optimization for semantic video segmentation.
\newblock In {\em CVPR}, 2016.

\bibitem{lee2016multi}
B.~Lee, E.~Erdenee, S.~Jin, and P.~K. Rhee.
\newblock Multi-class multi-object tracking using changing point detection.
\newblock {\em arXiv preprint}, 2016.

\bibitem{liu2008siftflow}
C.~Liu, J.~Yuen, A.~Torralba, J.~Sivic, and W.~T. Freeman.
\newblock Sift flow: dense correspondence across difference scenes.
\newblock In {\em ECCV}, 2008.

\bibitem{liu2016ssd}
W.~Liu, D.~Anguelov, D.~Erhan, C.~Szegedy, and S.~Reed.
\newblock Ssd: Single shot multibox detector.
\newblock 2016.

\bibitem{long2015fully}
J.~Long, E.~Shelhamer, and T.~Darrell.
\newblock Fully convolutional networks for semantic segmentation.
\newblock In {\em CVPR}, 2015.

\bibitem{silberman2012indoor}
P.~K. Nathan~Silberman, Derek~Hoiem and R.~Fergus.
\newblock Indoor segmentation and support inference from rgbd images.
\newblock In {\em ECCV}, 2012.

\bibitem{ouyang2015deepid}
W.~Ouyang, X.~Wang, X.~Zeng, S.~Qiu, P.~Luo, Y.~Tian, H.~Li, S.~Yang, Z.~Wang,
  and C.-C. Loy.
\newblock Deepid-net: Deformable deep convolutional neural networks for object
  detection.
\newblock In {\em CVPR}, 2015.

\bibitem{patraucean2015spatio}
V.~Patraucean, A.~Handa, and R.~Cipolla.
\newblock Spatio-temporal video autoencoder with differentiable memory.
\newblock {\em arXiv preprint arXiv:1511.06309}, 2015.

\bibitem{pfister2015flowing}
T.~Pfister, J.~Charles, and A.~Zisserman.
\newblock Flowing convnets for human pose estimation in videos.
\newblock In {\em ICCV}, 2015.

\bibitem{ranjan2016optical}
A.~Ranjan and M.~J. Black.
\newblock Optical flow estimation using a spatial pyramid network.
\newblock {\em arXiv preprint}, 2016.

\bibitem{rastegari2016xnor}
M.~Rastegari, V.~Ordonez, J.~Redmon, and A.~Farhadi.
\newblock Xnor-net: Imagenet classification using binary convolutional neural
  networks.
\newblock {\em arXiv preprint}, 2016.

\bibitem{ren2015faster}
S.~Ren, K.~He, R.~Girshick, and J.~Sun.
\newblock {Faster R-CNN: Towards real-time object detection with region
  proposal networks}.
\newblock In {\em NIPS}, 2015.

\bibitem{revaud2015epicflow}
J.~Revaud, P.~Weinzaepfel, Z.~Harchaoui, and C.~Schmid.
\newblock {EpicFlow: Edge-Preserving Interpolation of Correspondences for
  Optical Flow}.
\newblock In {\em CVPR}, 2015.

\bibitem{olga2015imagenet}
O.~Russakovsky, J.~Deng, H.~Su, J.~Krause, S.~Satheesh, S.~Ma, Z.~Huang,
  A.~Karpathy, A.~Khosla, M.~Bernstein, A.~C. Berg, and L.~Fei-Fei.
\newblock {ImageNet Large Scale Visual Recognition Challenge}.
\newblock {\em IJCV}, 2015.

\bibitem{laura2016optical}
L.~Sevilla-Lara, D.~Sun, V.~Jampani, and M.~J. Black.
\newblock Optical flow with semantic segmentation and localized layers.
\newblock In {\em CVPR}, 2016.

\bibitem{shelhamer2016clockwork}
E.~Shelhamer, K.~Rakelly, J.~Hoffman, and T.~Darrell.
\newblock Clockwork convnets for video semantic segmentation.
\newblock In {\em ECCV}, 2016.

\bibitem{simonyan2015very}
K.~Simonyan and A.~Zisserman.
\newblock Very deep convolutional networks for large-scale image recognition.
\newblock In {\em ICLR}, 2015.

\bibitem{sun2014dl}
L.~Sun, K.~Jia, T.-H. Chan, Y.~Fang, G.~Wang, and S.~Yan.
\newblock Dl-sfa: deeply-learned slow feature analysis for action recognition.
\newblock In {\em CVPR}, 2014.

\bibitem{szegedy2015going}
C.~Szegedy, W.~Liu, Y.~Jia, P.~Sermanet, S.~Reed, D.~Anguelov, D.~Erhan,
  V.~Vanhoucke, and A.~Rabinovich.
\newblock Going deeper with convolutions.
\newblock In {\em CVPR}, 2015.

\bibitem{szegedy2016rethinking}
C.~Szegedy, V.~Vanhoucke, S.~Ioffe, J.~Shlens, and Z.~Wojna.
\newblock Rethinking the inception architecture for computer vision.
\newblock In {\em CVPR}, 2016.

\bibitem{weickert2006survey}
J.~Weickert, A.~Bruhn, T.~Brox, and N.~Papenberg.
\newblock A survey on variational optic flow methods for small displacements.
\newblock In {\em Mathematical models for registration and applications to
  medical imaging}. 2006.

\bibitem{weinzaepfel2013deepflow}
P.~Weinzaepfel, J.~Revaud, Z.~Harchaoui, and C.~Schmid.
\newblock {DeepFlow: Large displacement optical flow with deep matching}.
\newblock In {\em CVPR}, 2013.

\bibitem{wiskott2002slow}
L.~Wiskott and T.~J. Sejnowski.
\newblock Slow feature analysis: Unsupervised learning of invariances.
\newblock {\em Neural computation}, 2002.

\bibitem{yang2016craft}
B.~Yang, J.~Yan, Z.~Lei, and S.~Z. Li.
\newblock Craft objects from images.
\newblock In {\em CVPR}, 2016.

\bibitem{zeiler2014visualizing}
M.~D. Zeiler and R.~Fergus.
\newblock Visualizing and understanding convolutional networks.
\newblock In {\em ECCV}, 2014.

\bibitem{zhang2011discriminative}
W.~Zhang, P.~Srinivasan, and J.~Shi.
\newblock Discriminative image warping with attribute flow.
\newblock In {\em CVPR}, 2011.

\bibitem{zhang2015accelerating}
X.~Zhang, J.~Zou, K.~He, and J.~Sun.
\newblock Accelerating very deep convolutional networks for classification and
  detection.
\newblock {\em TPAMI}, 2015.

\bibitem{zhang2012slow}
Z.~Zhang and D.~Tao.
\newblock Slow feature analysis for human action recognition.
\newblock {\em TPAMI}, 2012.

\bibitem{zheng2015conditional}
S.~Zheng, S.~Jayasumana, B.~Romera-Paredes, V.~Vineet, Z.~Su, D.~Du, C.~Huang,
  and P.~Torr.
\newblock Conditional random fields as recurrent neural networks.
\newblock In {\em ICCV}, 2015.

\bibitem{zou2012deep}
W.~Zou, S.~Zhu, K.~Yu, and A.~Y. Ng.
\newblock Deep learning of invariant features via simulated fixations in video.
\newblock In {\em NIPS}, 2012.

\end{thebibliography}
}

\end{document}